\documentclass{article}

\usepackage{microtype}
\usepackage{graphicx}
\usepackage{subfigure}
\usepackage{booktabs} % for professional tables

\usepackage{hyperref}

\usepackage{algorithm}
\usepackage[noend]{algpseudocode}
\usepackage{bbm}
\usepackage{nicefrac}
\usepackage{amsmath,amsfonts,bm}
\usepackage{mathtools}

\def\eqref#1{equation~\ref{#1}}
\def\Eqref#1{Equation~\ref{#1}}
\def\1{\bm{1}}

\def\eps{{\epsilon}}

\DeclareMathAlphabet{\mathsfit}{\encodingdefault}{\sfdefault}{m}{sl}
\SetMathAlphabet{\mathsfit}{bold}{\encodingdefault}{\sfdefault}{bx}{n}

\def\gH{{\mathcal{H}}}

\DeclarePairedDelimiterX{\infdivx}[2]{(}{)}{%
  #1\;\delimsize\|\;#2%
}

\DeclareMathOperator*{\argmax}{arg\,max}
\DeclareMathOperator*{\argmin}{arg\,min}

\usepackage{xspace}
\usepackage[utf8]{inputenc}
\usepackage[english]{babel}
\usepackage{amsthm}
\usepackage{amssymb}
\usepackage{mathtools}
\usepackage{hyperref}
\usepackage{appendix}
\usepackage[normalem]{ulem}

\usepackage{etoolbox}
\makeatletter
\patchcmd{\hyper@makecurrent}{%
    \ifx\Hy@param\Hy@chapterstring
        \let\Hy@param\Hy@chapapp
    \fi
}{%
    \iftoggle{inappendix}{%true-branch
        \@checkappendixparam{chapter}%
        \@checkappendixparam{section}%
        \@checkappendixparam{subsection}%
        \@checkappendixparam{subsubsection}%
        \@checkappendixparam{paragraph}%
        \@checkappendixparam{subparagraph}%
    }{}%
}{}{\errmessage{failed to patch}}

\newcommand*{\@checkappendixparam}[1]{%
    \def\@checkappendixparamtmp{#1}%
    \ifx\Hy@param\@checkappendixparamtmp
        \let\Hy@param\Hy@appendixstring
    \fi
}
\makeatletter

\newtoggle{inappendix}
\togglefalse{inappendix}

\apptocmd{\appendix}{\toggletrue{inappendix}}{}{\errmessage{failed to patch}}
\apptocmd{\subappendices}{\toggletrue{inappendix}}{}{\errmessage{failed to patch}}

\newcommand\strike{\bgroup\markoverwith{\textcolor{red}{\rule[0.5ex]{2pt}{0.4pt}}}\ULon}
\newtheorem*{rep@theorem}{\rep@title}
\newcommand{\newreptheorem}[2]{%
\newenvironment{rep#1}[1]{%
 \def\rep@title{#2 \ref{##1}}%
 \begin{rep@theorem}}%
 {\end{rep@theorem}}}
\makeatother
\newreptheorem{theorem}{Theorem}
\newreptheorem{lemma}{Lemma}
\newreptheorem{assumption}{Assumption}

\newtheorem{theorem}{Theorem}[section]

\newtheorem{lemma}[theorem]{Lemma}

\usepackage{accents}
\usepackage{cancel}

\usepackage[accepted]{icml2019}

\icmltitlerunning{Maximum Entropy reinforcement learning with Mixture Policies}

\begin{document}

% \twocolumn[
\onecolumn
\icmltitle{Maximum Entropy Reinforcement Learning with Mixture Policies}

% It is OKAY to include author information, even for blind
% submissions: the style file will automatically remove it for you
% unless you've provided the [accepted] option to the icml2019
% package.

% List of affiliations: The first argument should be a (short)
% identifier you will use later to specify author affiliations
% Academic affiliations should list Department, University, City, Region, Country
% Industry affiliations should list Company, City, Region, Country

% You can specify symbols, otherwise they are numbered in order.
% Ideally, you should not use this facility. Affiliations will be numbered
% in order of appearance and this is the preferred way.
% \icmlsetsymbol{equal}{*}

\begin{icmlauthorlist}
\icmlauthor{Nir Baram}{to}
\icmlauthor{Guy Tennenholtz}{to}
\icmlauthor{Shie Mannor}{to}
\end{icmlauthorlist}

\icmlaffiliation{to}{Technion, Israel Institute of Technology}
\icmlcorrespondingauthor{Nir Baram}{bentzinir@gmail.com }

% You may provide any keywords that you
% find helpful for describing your paper; these are used to populate
% the "keywords" metadata in the PDF but will not be shown in the document
\icmlkeywords{Machine Learning, ICML}

\vskip 0.3in
% ]

% this must go after the closing bracket ] following \twocolumn[ ...

% This command actually creates the footnote in the first column
% listing the affiliations and the copyright notice.
% The command takes one argument, which is text to display at the start of the footnote.
% The \icmlEqualContribution command is standard text for equal contribution.
% Remove it (just {}) if you do not need this facility.

%\printAffiliationsAndNotice{}  % leave blank if no need to mention equal contribution
\printAffiliationsAndNotice{\icmlEqualContribution} % otherwise use the standard text.

\begin{abstract}
Mixture models are an expressive hypothesis class that can approximate a rich set of policies. However, using mixture policies in the Maximum Entropy (MaxEnt) framework is not straightforward. The entropy of a mixture model is not equal to the sum of its components, nor does it have a closed-form expression in most cases. Using such policies in MaxEnt algorithms, therefore, requires constructing a tractable approximation of the mixture entropy. In this paper, we derive a simple, low-variance mixture-entropy estimator. We show that it is closely related to the sum of marginal entropies. Equipped with our entropy estimator, we derive an algorithmic variant of Soft Actor-Critic (SAC) to the mixture policy case and evaluate it on a series of continuous control tasks.
\end{abstract}

\section{Introduction}
Maximum Entropy (MaxEnt) reinforcement learning algorithms \cite{ziebart2008maximum} are among the state-of-the-art methods in off-policy, continuous control reinforcement learning (RL) \cite{haarnoja2018soft, yin2002maximum, sac}. MaxEnt algorithms achieve superior performance by offering a structured approach to exploration. Particularly, they add the entropy over actions to the discounted sum of returns. This modified objective encourages policies to maximize the reward while being as erratic as possible, which highly assists with exploration and stability \cite{ziebart2010modeling}. Nevertheless, contemporary MaxEnt techniques use simple policy classes (e.g. Normal) whose entropy function has a closed-form expression.
\par
For the MaxEnt framework to remain competent in complex control tasks, it should be extended to handle richer classes of policies capable of accommodating complex behaviors. In this gap, much attention has been recently given to approaches that do not require knowledge of the probability of the performed action $a$ in a state $s$, $\pi(a | s)$ \cite{song2020optimistic}. For example, Distributional Policy Optimization (DPO) \cite{tessler2019distributional} builds upon the policy gradient framework \cite{kakade2001natural}, to propose a \emph{Generative Actor} whose optimization is based on Quantile Regression \cite{koenker2001quantile}. However, methods of this type usually involve challenging optimization, long training times, and lack the exploration and stability properties of MaxEnt algorithms.
\par
\begin{figure*}[t]\label{fig:gaussians}
\centering
\includegraphics[width=0.9\textwidth, height=0.2\textheight]{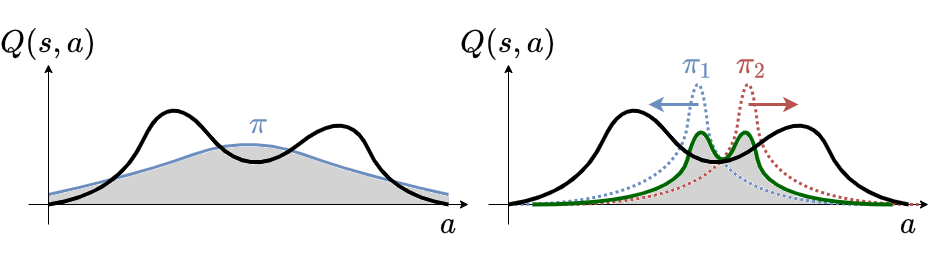}
\caption{\textbf{Illustration of Entropy vs Mixture Entropy regularization}. Complex control tasks are more likely to elicit non-unimodal value functions (black line). In such situations, unimodal parameterized policies (left) run the risk of converging to suboptimal local minima. To explore both extremes of $Q$, we need to allow high entropy (gray area under the curve) by forcing high entropy regularization. In contrast, a mixture model (right) readily fits both centers without suffering high entropy (gray area under the curve). In this case, each component in the mixture can \emph{latch} to a different extremum of $Q$. The resultant mixture policy is more diverse, and therefore robust since it assigns proportional weights to the two local maxima of $Q$.}
\end{figure*}

Instead, we propose to use mixture models \cite{agostini2010reinforcement} to construct an expressive policy. Mixture models are universal density estimators \cite{carreira2000mode}. They can approximate any distribution using elementary components with a closed-form probability density function (see Figure~\ref{fig:gaussians} for an illustration). Nevertheless, combining mixture policies with MaxEnt algorithms poses a challenge of estimating the entropy of the mixture, as the latter does not have a closed-form expression in many scenarios \cite{kim2015entropy}, nor is it easy to approximate \cite{huber2008entropy}.
\par
In this paper, we attempt to effectively integrate mixture policies with the MaxEnt framework. We derive a tractable, low-variance mixture-entropy estimator, reminiscent of the weighted sum of the marginal entropies of the mixing components. As such, it is intuitive to understand and straightforward to calculate. In a series of experiments we empirically demonstrate that soft actor-critic algorithms can optimize mixture models comparably well to single-component policies. Optimizing the mixing weights as well, which is outside the scope of this work, may result in an additional performance gain.
\par
The rest of the paper is organized as follows. Section~\ref{seq:preliminaries} outlines the basic mathematical background. Section~\ref{seq:maxent-entropy} presents our mixture MaxEnt low-variance estimator, followed by Section~\ref{seq:sacm} where we describe our Soft Actor-Critic variant with mixture policies. Section~\ref{seq:related-work} includes a short survey of mixture-models in RL and Section~\ref{seq:experiments} summarizes our empirical evaluation part. We conclude in Section~\ref{seq:conclusions}.

\section{Preliminaries}
\label{seq:preliminaries}
We assume a Markov Decision Process (MDP) ~\cite{puterman2014markov} which specifies the environment as a tuple $\mathcal{M} = \langle \mathcal{S},\mathcal{A}, r, P, \gamma, \rho_0 \rangle$, consisting of a state space $\mathcal{S}$, an action space $\mathcal{A}$, a reward function $r:\mathcal{S} \times \mathcal{A} \mapsto [0,1]$, a transition probability function $P:\mathcal{S} \times \mathcal{A}\times \mathcal{S} \mapsto [0,1]$, a discount factor $\gamma \in (0,1)$,  and an initial state distribution $\rho_0: \mathcal{S} \mapsto [0,1]$. A policy $\pi: \mathcal{S} \times \mathcal{A} \mapsto [0, 1]$ interacts with the environment sequentially, starting from an initial state ${s_0 \sim \rho_0}$.  At time $t=0,1,\cdots$ the policy produces a probability distribution over the action set $\mathcal{A}$ from which an action $a_t \in \mathcal{A}$ is sampled and played. The environment then generates a scalar reward $r(s_t, a_t)$ and a next state $s_{t+1}$ is sampled from the transition probability function $P(\cdot | s_t, a_t)$. 
% The value of a policy may be expressed by the state-action value function $Q^\pi(s,a)=\mathbb{E}^\pi \sum\limits_{t=0}^\infty \big[ \gamma^t r(s_t,a_t) | s_0=s, a_0=a \big]$,
% or the state value function $v^\pi(s)=\mathbb{E}_{a \sim \pi}Q^\pi(s,a)$.

We use ${\rho_\pi(s,a)\triangleq (1-\gamma) \sum\limits_{t=0}^\infty \gamma^t P^\pi(s_t=s, a_t=a | s_0 \sim \rho_0)}$ and $\rho_\pi(s)\triangleq \sum_a\ \rho_\pi(s,a)$ to denote the discounted state-action and state visitation distributions of policy $\pi$, respectively.

In this work, we assume a policy $\pi$ that is a mixture of $N$ components $\pi_i,...\pi_N$ with corresponding mixing weights $w_1, \cdots w_N$. We can treat the set of component weights as the probabilities that outcomes $1 \cdots N$ take place, given a random variable $W$, where $Pr(W = i) = w_i$.

\section{Maximum Mixture Entropy}
\label{seq:maxent-entropy}

MaxEnt methods promote stochastic policies by augmenting the standard expected sum of returns $
        J(\pi) = 
        \sum_{t=0}^T \underset{a_t,s_t \sim \rho_{\pi}}{\mathbb{E}}
        [r(s_t,a_t)$,
with the expected entropy of the policy over $\rho_\pi$, 
$\sum_{t=0}^T \mathbb{E}_{s_t,a_t \sim \rho_\pi}
        [r(s_t,a_t) + \alpha \gH(\pi(\cdot | s_t))]$.
Here, $\alpha > 0$ is the entropy coefficient and the entropy, $\gH(\pi(\cdot | s))$, is given by 
$
    \gH(\pi(\cdot|s)) = - \sum_a \pi(a|s)\log \pi(a|s)
$.
The parameter $\alpha$ controls the stochasticity of optimal policy by balancing the reward and entropy terms.
\subsection{Mixture Policies}
Various types of mixture models are discussed in the RL literature. In episodic-mixture methods, a component is selected at the start of each episode and plays for the entire episode. Such methods are sometimes referred to as Ensemble RL methods \cite{wiering2008ensemble}. A second kind, which we refer to as step-mixture methods, handpicks a new component to play after each or several steps. Step-mixtures describe popular RL setups such as \emph{macro actions} \cite{pickett2002policyblocks} or \emph{options} \cite{precup1998multi}. In a third setting, termed as action-mixture methods, actions are produced by combining the predictions of multiple models, e.g. as with combinatorial action spaces \cite{vinyals2019grandmaster}. In this work, we chiefly focus on the second type of step-mixture methods. The maximum entropy criterion in this setup is given by
\begin{align}\label{eq:maximum-mixture-entropy-objective}
        J(\pi) = 
        \sum_{t=0}^T \underset{\underset{a_t,s_t \sim \rho_{\pi_i}}{i \sim W}}{\mathbb{E}}&
        [r(s_t,a_t) + \alpha \gH(\pi(\cdot | s_t))].
\end{align}
Evaluating the objective in \Eqref{eq:maximum-mixture-entropy-objective} can be described by the following three steps sampling process: (1) sample a state from the mixture distribution $\rho_\pi$, (2) sample a mixture component $\pi_i$ with probability $w_i$, and (3) sample an action $a \sim \pi_i(\cdot | s_t)$ and evaluate. 

Notice that in Equation~\ref{eq:maximum-mixture-entropy-objective}, there exists a separate marginal state distribution $\rho_{\pi_i}$ for each mixture component $\pi_i$. As such, one can independently optimize the data term denoting the expected sum of returns as in the non-mixture case. Nevertheless, in neither the episodic nor the step-mixture settings can the second entropy term be separately optimized. The use of mixture entropy that depends on the overall mixture is therefore still a joint optimization problem.

\subsection{Mixture Entropy Estimation}
Mixture entropy regularization maximizes the expected entropy of a mixture policy $\gH(\pi)$. This is different from maximizing a weighted sum of marginal entropies $\sum_i w_i \gH(\pi_i)$. As we show below, the mixture entropy depends on the pairwise distances between the components. Therefore, maximizing the mixture entropy creates diversity between the components, which can provide substantial improvement in exploration and robustness \cite{ziebart2010modeling}. 

In a sequential decision-making setup, it is customary to define the entropy of $\pi$ as the expected entropy over $\rho_\pi$, i.e.,
\begin{align*}
        \gH(\pi) = \sum_{t=0}^T \mathbb{E}_{s_t \sim \rho_\pi}\gH(\cdot | s_t).
\end{align*}
Using laws of probability we can outline the connection between the joint and conditional mixture entropy:
\begin{align}\label{eq:entropy-identities}
        \gH(\pi,W) = \gH(\pi|W) + \gH(W) = \gH(W|\pi) + \gH(\pi),
\end{align}
where $\gH(\pi|W)$ is given by $\gH(\pi|W) = \underset{i}{\sum}w_i \gH(\pi_i)$, and $\gH(W|\pi)$ is defined as
\begin{align*}
 \gH(W|\pi) &= - \sum_{a \in A} \sum_{t=0}^T\mathbb{E}_{s_t \sim \rho_\pi}\pi(a|s_t)\gH(W|a, s_t)\\
& = - \sum_{t=0}^T\mathbb{E}_{s_t \sim \rho_\pi} \sum_{\underset{w \in W}{a \in A}} \pi(a|s_t) p(w|a,s_t) \log p(w|a,s_t)\\
& = - \sum_{t=0}^T\mathbb{E}_{s_t \sim \rho_\pi} \sum_{\underset{w \in W}{a \in A}} p(w,a|s_t) \log \frac{p(w,a|s_t)}{\pi(a|s_t)}.
\end{align*}

A closer look at \Eqref{eq:entropy-identities} reveals that the explicit form of the mixture entropy is non-trivial as it requires calculating $p(w, a|s)$, i.e., the joint action-component probability. Eliminating the need to calculate this quantity requires some form of approximation. One method to approximate the mixture entropy is using Monte Carlo (MC) sampling. While MC may yield an unbiased estimate it comes with a high computational cost. A different approach is to use analytic estimators to approximate the mixture entropy. Such estimators have estimation bias but are efficient to compute. 

A straightforward lower-bound estimate of the mixture entropy is given by
\begin{align}\label{eq:entropy-lower-bound}
    \gH(\pi) \geq \gH(\pi | W) = \underset{i}{\sum}w_i \gH(\pi_i),
\end{align} as conditioning can only decrease entropy.
Alternatively, an immediate upper bound is given by the joint entropy
\begin{align}\label{eq:entropy-upper-bound}
    \gH(\pi) \leq \gH(\pi,W)=\gH(\pi|W) + \gH(W)
\end{align}
The upper and lower bounds remain the same regardless of the ``overlap" between components. As such, they are poor choices for optimization problems that involve varying component locations.

Other estimators that depend on the "overlap" between components are based on Jensen's inequality \citep{cover1999elements} or kernel density estimation \cite{joe1989estimation, hall1993estimation}. However, these exhibit significant underestimation \cite{kolchinsky2017estimating}. To address this problem, \citet{kolchinsky2017estimating} proposed to estimate the mixture entropy using pairwise distances between mixture components. Specifically, let $D(p_i||p_j)$ denote a (generalized) distance function between probability measures $p_i$ and $p_j$. Whenever $D$ satisfies $D(p_i||p_j)=0$ for $p_i=p_j$\footnote{Note that we do not assume $D$ to obey the triangle inequality, nor that it is symmetric.}, then $\gH_D(\pi)$ can be used to approximate $\gH(\pi)$ by
\begin{align}\label{eq:entropy-pairwise-estimator}
    \gH_D(\pi) = \gH(\pi|W) -\underset{i}{\sum}w_i \log \underset{j}{\sum}w_j e^{-D(\pi_i||\pi_j)}.
\end{align}

\citet{kolchinsky2017estimating} also showed that for any distance function $D$, $\gH_D(\pi)$ is bounded by the joint and conditional entropy
\begin{align}\label{eq:pairwise-entropy-bounds}
    \gH(\pi|W) \leq \gH_D(\pi) \leq \gH(\pi,W).
\end{align}

Combining \Eqref{eq:pairwise-entropy-bounds} with \Eqref{eq:entropy-lower-bound} and \Eqref{eq:entropy-upper-bound} we can bound the bias of $\gH_D(\pi)$ by
\begin{align*}
    |\gH_D(\pi) - \gH(\pi)| \leq \gH(\pi,W) - \gH(\pi|W) = \gH(W).
\end{align*}
Tighter bounds can be achieved for specific distance functions. We refer the reader to \citet{kolchinsky2017estimating} for further information.

\subsection{Low Variance Estimation of $\gH_D(\pi)$}
$\gH_D(\pi)$ is attractive to use whenever $D$ is easy to approximate. The first term, corresponding to the marginal entropies, can be approximated at state $s$ by
\begin{align}\label{eq:marginal-entropies}
    \gH(\pi|W) = \sum_i w_i \gH(\pi_i) \approx - \sum_i w_i \log \pi_i (\tilde{a}_i | s),
\end{align}
where $\tilde{a}_i \sim \pi_i (\cdot | s)$. Setting $D$ to be the Kullback-Leibler divergence, the second term in $\gH_D(\pi)$ can be approximated at each state $s$ by
\begin{align}\nonumber
    & \underset{i}{\sum}w_i \log \underset{j}{\sum}w_j \exp (-D(\pi_i||\pi_j))
    \\\nonumber
    & = \underset{i}{\sum}w_i \log \underset{j}{\sum}w_j \exp \Big( {\mathbb{E}_{a\sim \pi_i} \log \frac{\pi_j(a|s)}{\pi_i (a|s)}} \Big)
    \\\nonumber
    & \approx \underset{i}{\sum}w_i \log \underset{j}{\sum}w_j \frac{\pi_j(\hat{a}_i|s)}{\pi_i (\hat{a}_i|s)}
    \\\nonumber
    & = \underset{i}{\sum}w_i \log \underset{j}{\sum}w_j \pi_j(\hat{a}_i|s) - \underset{i}{\sum}w_i \log \pi_i(\hat{a}_i|s),
\end{align}
where $\hat{a}_i \sim \pi_i (\cdot | s)$ is a second sample from policy $\pi_i$. Setting $\hat{a_i} \equiv \tilde{a}_i$, will result in a biased estimation of $\gH_D(\pi)$ since we use the same sample twice. However, the variance of the estimation will be significantly smaller as $\underset{i}{\sum}w_i \log \pi_i(\hat{a}_i|s)$ will cancel out from both terms (We refer the reader to Appendix~\ref{sec:unbiased-lower-bound} for an unbiased lower bound of $\hat{\gH}_D(\pi)$). Specifically, we have that
\begin{align}\nonumber
    \hat{\gH}&_D(\pi) \approx - \sum_i w_i \log \pi_i (\tilde{a}_i | s)
     - \Bigg[ \underset{i}{\sum}w_i \log \underset{j}{\sum}w_j \pi_j(\hat{a}_i|s) - \sum_i w_i \log \pi_i (\tilde{a}_i | s)\Bigg].
\end{align}
We thus have the following approximation for $\hat{\gH}_D(\pi)$
\begin{align}\label{eq:entropy-pairwise-estimator-simple}
    \hat{\gH}_D(\pi) \approx \underset{i}{\sum}w_i \hat{\gH}_i(\pi).
\end{align}
where $\hat{\gH}_i(\pi) = -\log \underset{j}{\sum}w_j \pi_j(\hat{a}_i|s).$

To see that $\text{Var}\big( \hat{\gH}_D(\pi) \big) \leq \text{Var}\big( \gH_D(\pi) \big)$ we denote by $X_i(s)$ the random variable $\log \pi_i(\cdot |s)$, and $Y_i(s)$ the random variable $\log \underset{j}{\sum}w_j \pi_j(\cdot |s)$. We have that
\begin{align*}
 \text{Var}\bigg(\gH_D(\pi) \bigg) &= 4 \bigg[ \text{Var}\bigg( \sum_i w_i X_i(s) \bigg) 
 +\text{Cov} \bigg( \sum_i w_i X_i(s), \sum_i w_i Y_i(s)\bigg) \bigg] 
+ \text{Var}\bigg( \sum_i w_i Y_i(s) \bigg)
\\
& \geq \text{Var}\bigg( \sum_i w_i Y_i(s) \bigg) = \text{Var}\bigg(\hat{\gH}_D(\pi) \bigg).
\end{align*}

Carefully examining \Eqref{eq:entropy-pairwise-estimator-simple} with relation to the expressions for marginal entropies in \Eqref{eq:marginal-entropies}, our result indicates that the mixture entropy is defined as a weighted sum of \emph{mixed} marginal entropies. That is, while the marginal entropy of each component, $\gH_i(\pi)$, is approximated by $-\log \pi_i(a|s)$, the \emph{mixed} marginal entropy, $\hat{\gH}_i(\pi)$, is approximated by 
$-\log \sum_j w_j \pi_j(\hat{a}_i|s).$ While the $i$-th \textbf{marginal entropy}, $\gH(\pi_i)$, measures the $\log$ probability of an action $a$ sampled from the $i$-th component, the $i$-th \textbf{mixed marginal entropy}, $\hat{\gH}_i(\pi)$, measures the $\log$ of the cumulative probability the mixture assigns to the same action. From an exploration viewpoint, the intrinsic bonus of any action will be \emph{silenced} if other mixture components are already exploring it.

\subsection{A Bellman Operator for Mixture Policies}
% It is possible to decompose the joint optimization function in the following way:
% \begin{align}\label{eq:decomposed-objective}
%     J(\pi) = \sum_{t=0}^T \mathbb{E}_{s_t,a_t \sim \rho_\pi}
%         r(s_t,a_t) + \alpha \sum_{i=1}^N w_i J_{\gH_i}(\pi),
% \end{align}
% where $J_{\gH_i}(\pi)$ is defined in the following way:
% \begin{align}
% J_{\gH_i}(\pi) = 
%         \sum_{t=0}^T \mathbb{E}_{s_t \sim \rho_\pi}
%         \hat{\gH}_i(\pi)(\cdot | s_t).
% \end{align}
% The decomposition in \Eqref{eq:decomposed-objective} allows us to learn separate value function, $v_{\gH_i}$, for each mixed marginal entropy, $\hat{\gH}_i(\pi)$. The Bellman backup operator of $v_{\gH_i}$ is defined as:
% \begin{align}\nonumber
%     \mathcal{T}^{\gH} v_{\gH_i}(s_t) \triangleq \mathbb{E}_{a_t \sim \pi_i} & - 
%     \log \underset{j}{\sum}w_j \pi_j(a_t|s_t) 
%     \\\nonumber
%     & + \mathbb{E}_{s_{t+1} \sim \rho} v_{\gH_i}(s_t).
% \end{align}
% This could prove useful for both variance reduction and algorithmic purposes, as shown in Section~\ref{seq:experiments}. 
In the infinite-horizon discounted setting we define the value function $v^\pi$ to include the entropy bonuses from every timestep
\begin{align*}
    v^{\pi}(s) =
    \underset{\tau \sim \pi}{\mathbb{E}}{ \bigg[ \sum_{t=0}^{\infty} \gamma^t \bigg( r(s_t, a_t) + \alpha \hat{\gH}_D(\pi(\cdot|s_t)) \bigg) \bigg| s_0 = s} \bigg],
\end{align*}
and similarly for the $Q$-function
$
    {Q^{\pi}(s,a) = \underset{\tau \sim \pi}{\mathbb{E}} \bigg[ \sum_{t=0}^{\infty} \gamma^t  r(s_t, a_t) 
    + \alpha \sum_{t=1}^{\infty} \gamma^t \hat{\gH}_D (\pi(\cdot|s_t)) \bigg| s_0 = s, a_0 = a \bigg].}
$
With this change, the soft value function can be written as
\begin{align}\label{eq:soft-value}
    v^{\pi}(s) = \underset{a \sim \pi}{\mathbb{E}}{Q^{\pi}(s,a)} + \alpha \hat{\gH}_D \left(\pi(\cdot|s)\right),
\end{align}
and the Bellman backup operator associated with the Q-function of the mixture policy is, therefore
${
    \mathcal{T}^\pi Q^\pi(s_t,a_t) \triangleq r(s_t,a_t) + \gamma \mathbb{E}_{s_{t+1} \sim \rho}v^\pi (s_{t+1})
}$.

\begin{algorithm*}[t!]
    % \hspace*{\algorithmicindent} 
	\caption{Soft Actor Critic Mixture (SACM)}
	\textbf{Initialization:} \\
	\hspace*{\algorithmicindent} $\theta, \{ \phi_i, \alpha_i\}_{i=0}^n$ \Comment{Initial parameters} \\
	\hspace*{\algorithmicindent} $ \bar{\theta} \leftarrow \theta$ \Comment{Initialize target network weights} \\
	\hspace*{\algorithmicindent} $ \mathcal{D} \leftarrow \emptyset$ \Comment{Initialize empty replay buffer}
	\begin{algorithmic}[1]
		\For {each environment step}
		    \State $i \sim [w_1,w_2,\cdots, w_N]$ \Comment{draw mixture component}
    		\State $a_t \sim \pi_i(\cdot|s_t)$ \Comment{Sample action from component $\pi_i$}
    		\State $s_{t+1} \sim p(s_{t+1}|s_t,a_t)$ \Comment{Sample transition from the environment}
    		\State $\mathcal{D} \leftarrow \mathcal{D} \bigcup \{ s_t, a_t, r_t, s_{t+1}\}$ \Comment{Store the transition in the replay buffer}
		\EndFor \State {\textbf{end for}}
		\\
		\For {each gradient step}
    		
    		\State $\theta \leftarrow \theta - \lambda_Q \nabla_{\theta} J_Q(\theta)$ \Comment{Update Q-function according to \eqref{eq:critic-gradient}}
        		
    		\State $\bar{\theta} \leftarrow \tau \theta + (1-\tau) \bar{\theta}$ \Comment{Polyak averaging of target Q function weights}
    		
		    \For {each mixture component $j$}
        		
        		\State $\phi_j \leftarrow \phi_j - \lambda_\pi \nabla_{\phi_j} J_\pi(\phi_j)$ \Comment{Update policy weights according to \eqref{eq:policy-gradient}}
        		
        		\State $\alpha_j \leftarrow \alpha_j - \lambda_\alpha \nabla_{\alpha_j} J_\alpha(\alpha_j)$ \Comment{Update entropy temperature according to \eqref{eq:alpha-obj}}

    		\EndFor \State {\textbf{end for}}
		\EndFor \State {\textbf{end for}}
	\end{algorithmic} \label{alg:sacm}
	\textbf{Output:} $ \{\phi_v \}_{i=1}^N$ \Comment{Optimal policy parameterization weights}
\end{algorithm*}

\section{Soft Actor Critic Mixture}
\label{seq:sacm}
A natural algorithmic choice to optimize MaxEnt mixtures is by Soft Actor-Critic (SAC) algorithms \cite{haarnoja2018soft}. In this section we will describe Soft Actor Critic Mixture (SACM), our proposed framework.  We use neural network function approximation for both the value function and the mixture policy, and alternate between policy evaluation steps (optimizing Q), and policy improvement steps (optimizing $\pi$).

Consider a parameterized soft Q-function $Q_{\theta}(s,a)$ and a set of parameterized policies $\{ \pi_{\phi_i} \}$, e.g., a set of multivariate Gaussian distributions. We denote by $\theta$ the soft $Q$-function neural network parameters, and by by $\phi =\{ \phi_i\}_{i=1}^N$ the parameters of the policy networks\footnote{It is common practice to represent $\{ \pi_{\phi_i} \}$ using distinct ``heads" over a shared feature extractor.}. In what follows, we will derive update rules for $\theta$ and $\phi$.

\subsection{Soft Mixture Policy Evaluation}
The soft Q-function parameters are trained to minimizes the Bellman error
\begin{align*}
    & J_Q(\theta) = 
    \underset{(s_t,a_t) \sim \mathcal{D}}{\mathbb{E}} \Big[ \frac{1}{2} (Q_{\theta}(s_t,a_t) - (r_{s_t,a_t} + \gamma \underset{s_{t+1} \sim \rho}{\mathbb{E}} v_{\bar{\theta}}(s_{t+1}))^2 \Big],
\end{align*}
where $\bar{\theta}$ is a Polyak averaging (target network) of $\theta$ which has been shown to stabilize training \cite{mnih2015human}. Note that $v_{\theta}$, given by \Eqref{eq:soft-value}, is not parameterized separately. The stochastic gradient of $J_Q(\theta)$ is thus
\begin{align}\label{eq:critic-gradient}
    \nabla_{\theta} J_Q(\theta) = 
 \nabla_{\theta} Q_{\theta}(s_t,a_t) (Q_{\theta}(s_t,a_t) -
    y).
\end{align}
The target value $y$ is computed by sampling an action $a_{t+1}$ from the mixture policy in the next state $s_{t+1}$
\begin{align*}
    y = r_{s_t,a_t} & + \gamma Q_{\bar{\theta}}(s_{t+1}, a_{t+1})
    -\alpha \log \underset{j}{\sum}w_j \pi_j(a_{t+1}|s_{t+1}).
\end{align*}
\subsection{Soft Mixture Policy Improvement}
Following \citep{haarnoja2018soft, sac}, the policy parameters are trained to minimize the expected Kullback-Liebler with respect to the soft $Q$ mixture function
\begin{align}\label{eq:policy-kl}
    \pi_i \leftarrow \underset{\pi' \in \Pi}{\argmin}D_{KL}\Bigg( \pi'(\cdot | s_t) \Bigg|\Bigg| \frac{\exp (\frac{1}{\alpha}Q(s_t,\cdot))}{Z(s_t)} \Bigg),
\end{align}
where $Z$ is an intractable partition function needed to normalize the right-hand side of the KL divergence to a valid probability distribution. Fortunately, it does not contribute to the gradient and can thus be ignored. \Eqref{eq:policy-kl}, however, shows that all mixture components follow the same Boltzmann distribution induced by the shared soft Q-function. Therefore, without optimizing the mixing weights, we run the risk of having the mixture components converge to the same policy. As an intermediate solution, one may maintain an array of $N$ soft Q-functions, $\{Q_{\theta_i}\}_{i=1}^N$, one for each component, and train them using non-overlapping bootstrapped data samples. Exposing components to distinct data samples can assist in breaking the symmetry. \cite{osband2016deep}. The update rule in \Eqref{eq:policy-kl} is achieved by minimizing the following objective
\begin{align}\nonumber
    J_\pi(\phi_i) = \underset{s_t \sim \mathcal{D}}{\mathbb{E}} [ \underset{a_t \sim \pi_{\phi_i}}{\mathbb{E}} (\alpha \log \pi_{\phi_i}(a_t|s_t) - Q_{\theta}(s_t,a_t))].
\end{align}
Unfortunately, the above objective is non differnetiable, as it involves sampling from a probability distribution defined by $\pi_{\phi_i}$. Approximating the gradient is possible by using the likelihood ratio gradient estimator (REINFORCE, \cite{williams1992simple}) or using the reparametrization trick \cite{kingma2013auto, rezende2014stochastic}. Using the latter approach, actions are sampled according to
\begin{align}\nonumber
 \bm{a_t}=\mu_{\phi_i}(s_t) + \eps_t \cdot \sigma_{\phi_i}(s_t),
\end{align}
where $\eps_t \sim \mathcal{N}(0,1)$ is a sample from a spherical Gaussian.
The modified objective now becomes
\begin{align}\nonumber
    J_\pi(\phi_i) = \underset{\underset{\eps_t \sim \mathcal{N}(0,1)}{s_t \sim \mathcal{D}}}{\mathbb{E}} [ (\alpha \log \pi_{\phi_i}(\bm{a_t}|s_t) - Q_{\theta}(s_t,\bm{a_t}))],
\end{align}
and the stochastic gradient is given by
\begin{align}\label{eq:policy-gradient}
    \nabla_{\phi_i} & J_\pi(\phi_i) = \alpha \nabla_{\phi_i} \log \pi_{\phi_i}(\bm{a_t}|s_t) + 
    (\alpha \nabla_{\bm{a_t}} \log \pi_{\phi_i}(\bm{a_t}|s_t) - \nabla_{\bm{a_t}} Q(\bm{a_t}, s_t)) \nabla_{\phi_i}\bm{a_t}(\phi_i).
\end{align}
Finally, we optimize a distinct entropy coefficient $\alpha_i$ for each component in the mixture. The alpha parameter is adjusted such that a pre-defined target entropy rate is maintained. This is possible by alternating the minimization of the following criterion concurrently while optimizing the $Q$ function and the policy, i.e.,
\begin{align}\label{eq:alpha-obj}
    &J(\alpha_i) = \mathbb{E}_{a_t \sim \pi_i} [-\alpha_i \log \pi_i(a_t|s_t) - \alpha_i \bar{\gH}].
\end{align}
The final $\alpha$ value in \Eqref{eq:soft-value} is then given by
$
    \alpha = \sum_i w_i \alpha_i
$.
Pseudo code for the Soft Actor Critic Mixture (SACM) is outlined in Algorithm~\ref{alg:sacm}.

\section{Related Work}
\label{seq:related-work}
Mixture policies serve several policy optimization settings. In multi-objective reinforcement learning, where complementary goals are defined simultaneously, components of mixture policies are used to optimize different objectives \cite{vamplew2009constructing}. The authors in \citet{abbeel2004apprenticeship}, on the other hand, describe an apprenticeship learning algorithm that optimizes the feature expectation of a mixture policy to match that of an anonymous expert policy. In the same spirit, the algorithm presented in \citet{fu2017learning} relies on the mixture distribution of expert demonstrations and an imitation policy to solve an inverse RL problem.

Mixture policies are useful in multi-agent problems as well. \citet{lanctot2017unified} describes a multi-agent RL algorithm where the strategy of each player is the best response to a mixture policy, defined by the rest of the players. On the other hand, the work of \citet{czarnecki2018mix} describes \emph{Mix \& Match} as a curriculum learning framework that defines a mixture of policies of increasing-learning complexity, while \citet{henderson2017optiongan} uses a mixture policy to extract options. They view the \emph{policy-over-options}, i.e., the agent that gates between options, as a single-step mixing policy. There is also the work of \citet{calinon2013compliant} that defined a multi-objective criterion that is optimized using a Gaussian mixture model - one of the most popular mixture models in RL literature. Finally, \citet{agarwal2019striving} that presented \emph{Random Ensemble Mixture} (REM), a Q-learning algorithm that enforces Bellman consistency on random convex combinations of multiple Q-value estimates.
\par
Our work is also related to ensemble learning. In this context, a notable example is \emph{Bootstrapped DQN} \cite{osband2016deep}, which carries out temporally-extended exploration using a multi-head Q network architecture where each head trains on a distinct subset of the data. The ideas presented in \citet{osband2016deep} were later on formalized by \citet{lee2020sunrise} as part of the SUNRISE algorithm (Simple UNified framework for reinforcement learning using enSEmbles). SUNRISE is a three-ingredient framework: a) Bootstrap with random initializations, b) Weighted Bellman backup, and c) UCB exploration.

Ensemble methods are also a powerful "go-to" strategy for mitigating optimization hazards. \emph{Averaged}-DQN shows how to use ensembles for stabilizing the optimization of Q learning algorithms. They average an ensemble of target networks to reduce approximation error variance \cite{anschel2017averaged}. In \citet{fausser2011ensemble}, the authors combine the predictions of multiple agents to learn a robust joint state-value function, while in \citet{marivate2013ensemble} the authors show that better performance is possible when learning a weighted linear combination of Q-values from an ensemble of independent algorithms.
% \vspace{0.7cm}

\begin{figure*}[t!]
\centering     %%% not \center
\subfigure[Hopper-v2]{\label{fig:a}\includegraphics[width=0.24\linewidth]{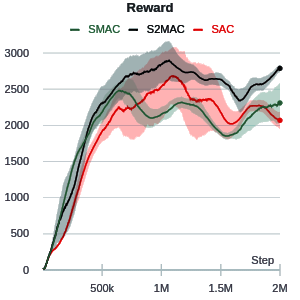}}
\subfigure[HalfCheetah-v2]{\label{fig:b}\includegraphics[width=0.24\linewidth]{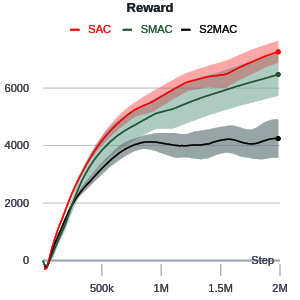}}
\subfigure[Ant-v2]{\label{fig:c}\includegraphics[width=0.24\linewidth]{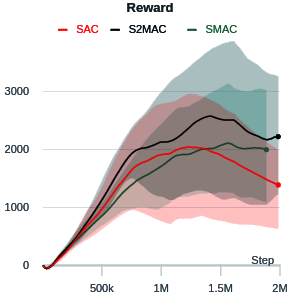}}
\\
\subfigure[Swimmer-v2]{\label{fig:d}\includegraphics[width=0.24\linewidth]{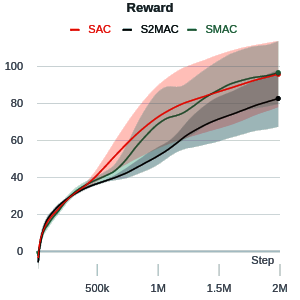}}
\subfigure[Walker-v2]{\label{fig:e}\includegraphics[width=0.24\linewidth]{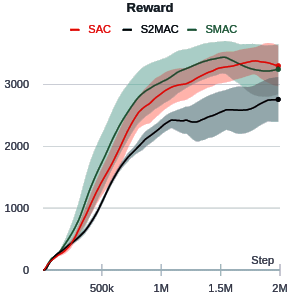}}
\subfigure[Humanoid-v2]{\label{fig:f}\includegraphics[width=0.24\linewidth]{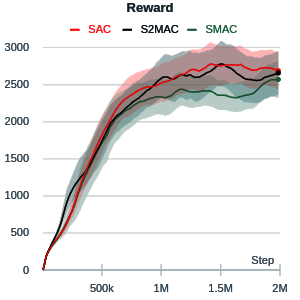}}
\caption{\textbf{Training curves for the continuous control benchmarks}. We use a mixture of $n=3$ components and report the best one. We evaluate both the SMAC and \emph{Semi}-SMAC (S2MAC) variants for the mixture model approach against the Soft Actor-Critic (SAC) baseline. The target entropy is set to $-\log |A|$ where $|A|$ is the size of the action vector. The results show the performance of stochastic action sampling from $\pi$ and not the behavior of playing the deterministic actions.}
\label{fig:results}
\end{figure*}

\section{Experiments}
\label{seq:experiments}
As the mixing weights are not optimized, we anticipate all components to collapse toward some mean policy $\bar{\pi}$. Consequently, the mixture entropy will equal the marginal entropy of the mean policy $\bar{\pi}$
\begin{align}\nonumber
    \hat{\gH}_D(\pi) & = - \underset{i}{\sum}w_i \log \underset{j}{\sum}w_j \pi_j(a_i|s)
 = - \underset{i}{\sum}w_i \log \bar{\pi}(\bar{a}|s) = - \log \bar{\pi}(\bar{a}|s),
\end{align}
where $\bar{a}$ is a sample from $\bar{\pi}$ and we use the fact that $\sum_j w_j=1$.
Since there is no reason to believe that mixture policies outperform single component MaxEnt policies under such conditions, we ask the following questions: (1) Do MaxEnt mixture policies perform \textbf{as well as} single-component MaxEnt policies? (2) Can performance improve by \textbf{enforcing asymmetry} in the entropy, even if the components' weights stay the same?

To answer the latter, we suggest to use a distinct soft Q-function, $Q_i$, for each component and a modified mixture entropy term
\begin{align}\label{eq:one-sided-mixture-ent}
    \hat{\gH}_D^{i}(\pi) & = - \underset{j\leq i}{\sum} \hat{\gH}_i(\pi).
\end{align}
According to \Eqref{eq:one-sided-mixture-ent}, the soft Q-function $Q_i$ is regulated by a partial sum of the marginal mixture entopies. By doing so, we aim to increase diversity between the Q functions as well as policy components. We call this variant Semi-Soft Actor-Critic Mixture (S2ACM).

\begin{figure}[!t]
\centering     %%% not \center
\subfigure[Varying Mixture Size]{\label{fig:g}\includegraphics[width=0.5\textwidth]{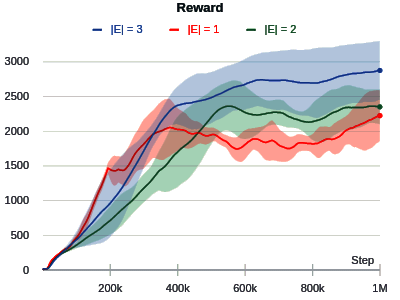}}
\caption{\textbf{Training Curves for Variable Mixture Sizes}. We run the S2MAC algorithm for 1 million steps on the Hopper-v2 task from the Mujoco control suite. The graph illustrates the training curves of mixtures of different sizes: $n=1,2,3$. The results of this experiment indicate that the algorithm's performance varies with mixture size. We attribute this finding to enhanced stability typically associated with ensemble methods.}
\label{fig:varying-mixture-size}
\end{figure}

\textbf{Testing Environment.} To evaluate our approach, we applied SACM on a variety of continuous control tasks in the MuJoCo physics simulator \cite{todorov2012mujoco}. Agents are comprised of linked bodies with distinct topologies. Some topologies are simple (Swimmer that contains 3 bodies) while others are complex (Humanoid that contains 17 bodies). The agent is rewarded for moving forward as much as possible in a given time frame. The state includes the positions and velocities of all bodies. The action is a vector of $n$ dimensions that defines the amount of torque applied to each joint.

\textbf{Implementation Details.} A neural network was used to implement the mixture policy with two fully-connected layers of width $300$ and $400$ and a ReLU activation after each layer. Each policy component, modeled as a multivariate Gaussian distribution, produced two outputs, representing the mean and the standard deviation. We experimented varying sizes of mixtures and report the number of components in our results. The critic accepted a state-action pair as input. The output layer of the critic, with $N$ neurons, represents the Q value of all components.

We ran each task for 3 million steps and report an average over 5 seeds. We used the Adam optimizer \cite{kingma2014adam} with a learning rate of $0.0003$ and a batch size of 256 to train the SACM algorithm. The algorithm collects experience for 1000 steps, followed by 1000 consecutive stochastic gradient steps. All other hyper-parameters are set to their default values as implemented in the Soft Actor-Critic algorithm as part of the stable-baselines framework \cite{stable-baselines}. The results are presented in Figure~\ref{fig:results}. We also include a comparison to relevant baselines, including (1) SAC \cite{sac}, (2) DDPG \cite{lillicrap2015continuous}, and (3) TD3 \cite{fujimoto2018addressing}.

\textbf{Mixture Size Sweep.}
We tested the effect of different mixture sizes on overall performance. We ran S2MAC on the Hopper-v2 control task with different values of $N$. Figure~\ref{fig:varying-mixture-size} shows an improvement when using higher-order mixtures.

\section{Conclusion}
\label{seq:conclusions}
Solving continuous control tasks with arbitrary policy classes presents the challenge of calculating the $\log$ probability of actions. Policy gradient methods use it ``as-is" to optimize the policy while MaxEnt algorithms require it to administer entropy regularization. To avoid using complicated techniques recently proposed to meet this challenge, we advocate the use of mixture policies.
\par
This paper extends the MaxEnt framework to handle mixture-policies, a class of universal density estimators. The novelty we present lies in the derivation of a tractable and intuitive mixture entropy estimator. The estimation we propose relies on a set of mixed marginal entropies. Each mixed-marginal, $\bar{\gH}_i$, measures the cumulative probability that all components assign to a sample from $\pi_i$. This stands in contrast to the case where the entropy of the $i$-th component is evaluated exclusively from $\pi_i$. From an exploration viewpoint, with this change, the intrinsic reward of an action sampled from mixture-component $\pi_i$ will be low if other mixture components are already exploring it. Our estimator allows MaxEnt algorithms to be applied to a broader class of policies, making them competitive with recent policy gradient approaches capable of optimizing non-unimodal policies. 
\par
Our empirical testing demonstrates that mixture policies match the state-of-the-art results of continuous control algorithms. However, we did not find clear evidence that mixture policies outperform their single-component counterparts. We attribute this finding to two reasons. First, benchmarked tasks are unimodal in nature, so a unimodal policy should do. Second, that the mixture policy collapses to a mean policy in the presence of equal mixing weights. Confirming the first hypothesis requires testing our algorithm in more complex control tasks or alternatively designing tasks with a controlled number of modes in the optimal Q function. Regarding the second hypothesis, it is possible to optimize the weights as part of the Hierarchical reinforcement learning framework \cite{dietterich1998maxq}, or by using policy gradient for example. Future research will have to explore this.

% \newpage
% \clearpage
\bibliography{smac}

\begin{thebibliography}{45}
\providecommand{\natexlab}[1]{#1}
\providecommand{\url}[1]{\texttt{#1}}
\expandafter\ifx\csname urlstyle\endcsname\relax
  \providecommand{\doi}[1]{doi: #1}\else
  \providecommand{\doi}{doi: \begingroup \urlstyle{rm}\Url}\fi

\bibitem[Abbeel \& Ng(2004)Abbeel and Ng]{abbeel2004apprenticeship}
Abbeel, P. and Ng, A.~Y.
\newblock Apprenticeship learning via inverse reinforcement learning.
\newblock In \emph{Proceedings of the twenty-first international conference on
  Machine learning}, pp.\ ~1, 2004.

\bibitem[Agarwal et~al.(2019)Agarwal, Schuurmans, and
  Norouzi]{agarwal2019striving}
Agarwal, R., Schuurmans, D., and Norouzi, M.
\newblock Striving for simplicity in off-policy deep reinforcement learning.
\newblock \emph{arXiv preprint arXiv:1907.04543}, 2019.

\bibitem[Agostini \& Celaya(2010)Agostini and
  Celaya]{agostini2010reinforcement}
Agostini, A. and Celaya, E.
\newblock Reinforcement learning with a gaussian mixture model.
\newblock In \emph{The 2010 International Joint Conference on Neural Networks
  (IJCNN)}, pp.\  1--8. IEEE, 2010.

\bibitem[Anschel et~al.(2017)Anschel, Baram, and Shimkin]{anschel2017averaged}
Anschel, O., Baram, N., and Shimkin, N.
\newblock Averaged-dqn: Variance reduction and stabilization for deep
  reinforcement learning.
\newblock In \emph{International Conference on Machine Learning}, pp.\
  176--185. PMLR, 2017.

\bibitem[Calinon et~al.(2013)Calinon, Kormushev, and
  Caldwell]{calinon2013compliant}
Calinon, S., Kormushev, P., and Caldwell, D.~G.
\newblock Compliant skills acquisition and multi-optima policy search with
  em-based reinforcement learning.
\newblock \emph{Robotics and Autonomous Systems}, 61\penalty0 (4):\penalty0
  369--379, 2013.

\bibitem[Carreira-Perpinan(2000)]{carreira2000mode}
Carreira-Perpinan, M.~A.
\newblock Mode-finding for mixtures of gaussian distributions.
\newblock \emph{IEEE Transactions on Pattern Analysis and Machine
  Intelligence}, 22\penalty0 (11):\penalty0 1318--1323, 2000.

\bibitem[Cover(1999)]{cover1999elements}
Cover, T.~M.
\newblock \emph{Elements of information theory}.
\newblock John Wiley \& Sons, 1999.

\bibitem[Czarnecki et~al.(2018)Czarnecki, Jayakumar, Jaderberg, Hasenclever,
  Teh, Osindero, Heess, and Pascanu]{czarnecki2018mix}
Czarnecki, W.~M., Jayakumar, S.~M., Jaderberg, M., Hasenclever, L., Teh, Y.~W.,
  Osindero, S., Heess, N., and Pascanu, R.
\newblock Mix\&match-agent curricula for reinforcement learning.
\newblock \emph{arXiv preprint arXiv:1806.01780}, 2018.

\bibitem[Dietterich(1998)]{dietterich1998maxq}
Dietterich, T.~G.
\newblock The maxq method for hierarchical reinforcement learning.
\newblock In \emph{ICML}, volume~98, pp.\  118--126. Citeseer, 1998.

\bibitem[Fau{\ss}er \& Schwenker(2011)Fau{\ss}er and
  Schwenker]{fausser2011ensemble}
Fau{\ss}er, S. and Schwenker, F.
\newblock Ensemble methods for reinforcement learning with function
  approximation.
\newblock In \emph{International Workshop on Multiple Classifier Systems}, pp.\
   56--65. Springer, 2011.

\bibitem[Fu et~al.(2017)Fu, Luo, and Levine]{fu2017learning}
Fu, J., Luo, K., and Levine, S.
\newblock Learning robust rewards with adversarial inverse reinforcement
  learning.
\newblock \emph{arXiv preprint arXiv:1710.11248}, 2017.

\bibitem[Fujimoto et~al.(2018)Fujimoto, Van~Hoof, and
  Meger]{fujimoto2018addressing}
Fujimoto, S., Van~Hoof, H., and Meger, D.
\newblock Addressing function approximation error in actor-critic methods.
\newblock \emph{arXiv preprint arXiv:1802.09477}, 2018.

\bibitem[Haarnoja et~al.(2018{\natexlab{a}})Haarnoja, Zhou, Abbeel, and
  Levine]{sac}
Haarnoja, T., Zhou, A., Abbeel, P., and Levine, S.
\newblock Soft actor-critic: Off-policy maximum entropy deep reinforcement
  learning with a stochastic actor.
\newblock \emph{arXiv preprint arXiv:1801.01290}, 2018{\natexlab{a}}.

\bibitem[Haarnoja et~al.(2018{\natexlab{b}})Haarnoja, Zhou, Hartikainen,
  Tucker, Ha, Tan, Kumar, Zhu, Gupta, Abbeel, et~al.]{haarnoja2018soft}
Haarnoja, T., Zhou, A., Hartikainen, K., Tucker, G., Ha, S., Tan, J., Kumar,
  V., Zhu, H., Gupta, A., Abbeel, P., et~al.
\newblock Soft actor-critic algorithms and applications.
\newblock \emph{arXiv preprint arXiv:1812.05905}, 2018{\natexlab{b}}.

\bibitem[Hall \& Morton(1993)Hall and Morton]{hall1993estimation}
Hall, P. and Morton, S.~C.
\newblock On the estimation of entropy.
\newblock \emph{Annals of the Institute of Statistical Mathematics},
  45\penalty0 (1):\penalty0 69--88, 1993.

\bibitem[Henderson et~al.(2017)Henderson, Chang, Bacon, Meger, Pineau, and
  Precup]{henderson2017optiongan}
Henderson, P., Chang, W.-D., Bacon, P.-L., Meger, D., Pineau, J., and Precup,
  D.
\newblock Optiongan: Learning joint reward-policy options using generative
  adversarial inverse reinforcement learning.
\newblock \emph{arXiv preprint arXiv:1709.06683}, 2017.

\bibitem[Hill et~al.(2018)Hill, Raffin, Ernestus, Gleave, Kanervisto, Traore,
  Dhariwal, Hesse, Klimov, Nichol, Plappert, Radford, Schulman, Sidor, and
  Wu]{stable-baselines}
Hill, A., Raffin, A., Ernestus, M., Gleave, A., Kanervisto, A., Traore, R.,
  Dhariwal, P., Hesse, C., Klimov, O., Nichol, A., Plappert, M., Radford, A.,
  Schulman, J., Sidor, S., and Wu, Y.
\newblock Stable baselines.
\newblock \url{https://github.com/hill-a/stable-baselines}, 2018.

\bibitem[Huber et~al.(2008)Huber, Bailey, Durrant-Whyte, and
  Hanebeck]{huber2008entropy}
Huber, M.~F., Bailey, T., Durrant-Whyte, H., and Hanebeck, U.~D.
\newblock On entropy approximation for gaussian mixture random vectors.
\newblock In \emph{2008 IEEE International Conference on Multisensor Fusion and
  Integration for Intelligent Systems}, pp.\  181--188. IEEE, 2008.

\bibitem[Joe(1989)]{joe1989estimation}
Joe, H.
\newblock Estimation of entropy and other functionals of a multivariate
  density.
\newblock \emph{Annals of the Institute of Statistical Mathematics},
  41\penalty0 (4):\penalty0 683--697, 1989.

\bibitem[Kakade(2001)]{kakade2001natural}
Kakade, S.~M.
\newblock A natural policy gradient.
\newblock \emph{Advances in neural information processing systems},
  14:\penalty0 1531--1538, 2001.

\bibitem[Kim et~al.(2015)Kim, Do, Oechtering, and Peters]{kim2015entropy}
Kim, S.~M., Do, T.~T., Oechtering, T.~J., and Peters, G.
\newblock On the entropy computation of large complex gaussian mixture
  distributions.
\newblock \emph{IEEE Transactions on Signal Processing}, 63\penalty0
  (17):\penalty0 4710--4723, 2015.

\bibitem[Kingma \& Ba(2014)Kingma and Ba]{kingma2014adam}
Kingma, D.~P. and Ba, J.
\newblock Adam: A method for stochastic optimization.
\newblock \emph{arXiv preprint arXiv:1412.6980}, 2014.

\bibitem[Kingma \& Welling(2013)Kingma and Welling]{kingma2013auto}
Kingma, D.~P. and Welling, M.
\newblock Auto-encoding variational bayes.
\newblock \emph{arXiv preprint arXiv:1312.6114}, 2013.

\bibitem[Koenker \& Hallock(2001)Koenker and Hallock]{koenker2001quantile}
Koenker, R. and Hallock, K.~F.
\newblock Quantile regression.
\newblock \emph{Journal of economic perspectives}, 15\penalty0 (4):\penalty0
  143--156, 2001.

\bibitem[Kolchinsky \& Tracey(2017)Kolchinsky and
  Tracey]{kolchinsky2017estimating}
Kolchinsky, A. and Tracey, B.~D.
\newblock Estimating mixture entropy with pairwise distances.
\newblock \emph{Entropy}, 19\penalty0 (7):\penalty0 361, 2017.

\bibitem[Lanctot et~al.(2017)Lanctot, Zambaldi, Gruslys, Lazaridou, Tuyls,
  P{\'e}rolat, Silver, and Graepel]{lanctot2017unified}
Lanctot, M., Zambaldi, V., Gruslys, A., Lazaridou, A., Tuyls, K., P{\'e}rolat,
  J., Silver, D., and Graepel, T.
\newblock A unified game-theoretic approach to multiagent reinforcement
  learning.
\newblock In \emph{Advances in neural information processing systems}, pp.\
  4190--4203, 2017.

\bibitem[Lee et~al.(2020)Lee, Laskin, Srinivas, and Abbeel]{lee2020sunrise}
Lee, K., Laskin, M., Srinivas, A., and Abbeel, P.
\newblock Sunrise: A simple unified framework for ensemble learning in deep
  reinforcement learning.
\newblock \emph{arXiv preprint arXiv:2007.04938}, 2020.

\bibitem[Lillicrap et~al.(2015)Lillicrap, Hunt, Pritzel, Heess, Erez, Tassa,
  Silver, and Wierstra]{lillicrap2015continuous}
Lillicrap, T.~P., Hunt, J.~J., Pritzel, A., Heess, N., Erez, T., Tassa, Y.,
  Silver, D., and Wierstra, D.
\newblock Continuous control with deep reinforcement learning.
\newblock \emph{arXiv preprint arXiv:1509.02971}, 2015.

\bibitem[Marivate \& Littman(2013)Marivate and Littman]{marivate2013ensemble}
Marivate, V. and Littman, M.
\newblock An ensemble of linearly combined reinforcement-learning agents.
\newblock In \emph{Proceedings of the 17th AAAI Conference on Late-Breaking
  Developments in the Field of Artificial Intelligence}, pp.\  77--79, 2013.

\bibitem[Mnih et~al.(2015)Mnih, Kavukcuoglu, Silver, Rusu, Veness, Bellemare,
  Graves, Riedmiller, Fidjeland, Ostrovski, et~al.]{mnih2015human}
Mnih, V., Kavukcuoglu, K., Silver, D., Rusu, A.~A., Veness, J., Bellemare,
  M.~G., Graves, A., Riedmiller, M., Fidjeland, A.~K., Ostrovski, G., et~al.
\newblock Human-level control through deep reinforcement learning.
\newblock \emph{nature}, 518\penalty0 (7540):\penalty0 529--533, 2015.

\bibitem[Osband et~al.(2016)Osband, Blundell, Pritzel, and
  Van~Roy]{osband2016deep}
Osband, I., Blundell, C., Pritzel, A., and Van~Roy, B.
\newblock Deep exploration via bootstrapped dqn.
\newblock In \emph{Advances in neural information processing systems}, pp.\
  4026--4034, 2016.

\bibitem[Pickett \& Barto(2002)Pickett and Barto]{pickett2002policyblocks}
Pickett, M. and Barto, A.~G.
\newblock Policyblocks: An algorithm for creating useful macro-actions in
  reinforcement learning.
\newblock In \emph{ICML}, volume~19, pp.\  506--513, 2002.

\bibitem[Precup \& Sutton(1998)Precup and Sutton]{precup1998multi}
Precup, D. and Sutton, R.~S.
\newblock Multi-time models for temporally abstract planning.
\newblock In \emph{Advances in neural information processing systems}, pp.\
  1050--1056, 1998.

\bibitem[Puterman(2014)]{puterman2014markov}
Puterman, M.~L.
\newblock \emph{Markov decision processes: discrete stochastic dynamic
  programming}.
\newblock John Wiley \& Sons, 2014.

\bibitem[Rezende et~al.(2014)Rezende, Mohamed, and
  Wierstra]{rezende2014stochastic}
Rezende, D.~J., Mohamed, S., and Wierstra, D.
\newblock Stochastic backpropagation and approximate inference in deep
  generative models.
\newblock \emph{arXiv preprint arXiv:1401.4082}, 2014.

\bibitem[Song \& Zhao(2020)Song and Zhao]{song2020optimistic}
Song, J. and Zhao, C.
\newblock Optimistic distributionally robust policy optimization.
\newblock \emph{arXiv preprint arXiv:2006.07815}, 2020.

\bibitem[Tessler et~al.(2019)Tessler, Tennenholtz, and
  Mannor]{tessler2019distributional}
Tessler, C., Tennenholtz, G., and Mannor, S.
\newblock Distributional policy optimization: An alternative approach for
  continuous control.
\newblock In \emph{Advances in Neural Information Processing Systems}, pp.\
  1352--1362, 2019.

\bibitem[Todorov et~al.(2012)Todorov, Erez, and Tassa]{todorov2012mujoco}
Todorov, E., Erez, T., and Tassa, Y.
\newblock Mujoco: A physics engine for model-based control.
\newblock In \emph{2012 IEEE/RSJ International Conference on Intelligent Robots
  and Systems}, pp.\  5026--5033. IEEE, 2012.

\bibitem[Vamplew et~al.(2009)Vamplew, Dazeley, Barker, and
  Kelarev]{vamplew2009constructing}
Vamplew, P., Dazeley, R., Barker, E., and Kelarev, A.
\newblock Constructing stochastic mixture policies for episodic multiobjective
  reinforcement learning tasks.
\newblock In \emph{Australasian joint conference on artificial intelligence},
  pp.\  340--349. Springer, 2009.

\bibitem[Vinyals et~al.(2019)Vinyals, Babuschkin, Czarnecki, Mathieu, Dudzik,
  Chung, Choi, Powell, Ewalds, Georgiev, et~al.]{vinyals2019grandmaster}
Vinyals, O., Babuschkin, I., Czarnecki, W.~M., Mathieu, M., Dudzik, A., Chung,
  J., Choi, D.~H., Powell, R., Ewalds, T., Georgiev, P., et~al.
\newblock Grandmaster level in starcraft ii using multi-agent reinforcement
  learning.
\newblock \emph{Nature}, 575\penalty0 (7782):\penalty0 350--354, 2019.

\bibitem[Wiering \& Van~Hasselt(2008)Wiering and
  Van~Hasselt]{wiering2008ensemble}
Wiering, M.~A. and Van~Hasselt, H.
\newblock Ensemble algorithms in reinforcement learning.
\newblock \emph{IEEE Transactions on Systems, Man, and Cybernetics, Part B
  (Cybernetics)}, 38\penalty0 (4):\penalty0 930--936, 2008.

\bibitem[Williams(1992)]{williams1992simple}
Williams, R.~J.
\newblock Simple statistical gradient-following algorithms for connectionist
  reinforcement learning.
\newblock \emph{Machine learning}, 8\penalty0 (3-4):\penalty0 229--256, 1992.

\bibitem[Yin(2002)]{yin2002maximum}
Yin, P.-Y.
\newblock Maximum entropy-based optimal threshold selection using deterministic
  reinforcement learning with controlled randomization.
\newblock \emph{Signal Processing}, 82\penalty0 (7):\penalty0 993--1006, 2002.

\bibitem[Ziebart(2010)]{ziebart2010modeling}
Ziebart, B.~D.
\newblock Modeling purposeful adaptive behavior with the principle of maximum
  causal entropy.
\newblock 2010.

\bibitem[Ziebart et~al.(2008)Ziebart, Maas, Bagnell, and
  Dey]{ziebart2008maximum}
Ziebart, B.~D., Maas, A.~L., Bagnell, J.~A., and Dey, A.~K.
\newblock Maximum entropy inverse reinforcement learning.
\newblock In \emph{Aaai}, volume~8, pp.\  1433--1438. Chicago, IL, USA, 2008.

\end{thebibliography}
\bibliographystyle{icml2019}

%%%%%%%%%%%%%%%%%%%%%%%%%%%%%%%%%%%%%%%%%%%%%%%%%%%%%%%%%%%%%%%%%%%%%%%%%%%%%%%
%%%%%%%%%%%%%%%%%%%%%%%%%%%%%%%%%%%%%%%%%%%%%%%%%%%%%%%%%%%%%%%%%%%%%%%%%%%%%%%
% DELETE THIS PART. DO NOT PLACE CONTENT AFTER THE REFERENCES!
%%%%%%%%%%%%%%%%%%%%%%%%%%%%%%%%%%%%%%%%%%%%%%%%%%%%%%%%%%%%%%%%%%%%%%%%%%%%%%%
%%%%%%%%%%%%%%%%%%%%%%%%%%%%%%%%%%%%%%%%%%%%%%%%%%%%%%%%%%%%%%%%%%%%%%%%%%%%%%%
\newpage
\appendix

\section{Mixture Entropy Lower Bound}
\label{sec:unbiased-lower-bound}
In the following we derive an unbiased lower bound for the mixture entropy. Denoting $z_i = \log \underset{j}{\sum}w_j\exp{(-D(\pi_i||\pi_j))}$, we can write \Eqref{eq:entropy-pairwise-estimator} more compactly as:
\begin{align}\label{eq:entropy-pairwise-estimator-compact}
    \hat{\gH}_D(\pi) = \gH(\pi|W) -\underset{i}{\sum}w_i z_i.
\end{align}

In the following, we wish to upper bound $z_i$ for all $i \in [N]$. For each $j \in [N]$ define $b_j=\frac{1}{N}$ and ${d_{i,j}=w_j \exp{(-D(\pi_i||\pi_j))}}$. Since $b_j \geq 0$ and $d_{i,j} \geq 0$ for all $j$, we can write the following inequality:
\begin{lemma}[Log Sum Inequality]
For non-negative numbers $d_{i,1},\cdots d_{i,N}$, and $b_1,...b_N$:
\begin{align}
    \sum_{j=1}^N a_j \log \frac{a_j}{b_j} \geq \Big( \sum_{j=1}^N a_j \Big) \log \frac{\sum_{j=1}^N a_j}{\sum_{j=1}^N b_j}
\end{align}
\end{lemma}

We have that:
\begin{align}
    & \log \sum_j a_j \leq \Big( \frac{1}{\sum_j a_j}\Big) \sum_j a_j \log \frac{a_j}{b_j}  \\
     & = \frac{1}{\sum_j a_j} \Big( \sum_j a_j \log a_j - a_j \log b_j \Big).
\end{align}

Since $0<b_j<1$ and $ 0< a_j < 1$ then $ {-C < a_j \log b_j < 0}$. Therefore:

\begin{align}
    & \frac{1}{\sum_j a_j} \Big( \sum_j a_j \log a_j - a_j \log b_j \Big) \\
    & \leq \frac{1}{\sum_j a_j} \Big( \sum_j a_j \log a_j + C \Big).
\end{align}

Next, we plug in the expression for $a_j = w_j \exp{(-D(\pi_i||\pi_j))}$. Since all mixing weights $w_i=w$ are equal we can further simplify:
\begin{align}
    & \frac{1}{\sum_j a_j} \Big( \sum_j a_j \log a_j + C \Big) \\
    & = \frac{1}{w\sum_j d_j} \Big( w\sum_j d_j \log w d_j + \frac{wC}{w} \Big)\\
    & = \frac{1}{\sum_j d_j} \Big( \sum_j d_j \log w  + d_j \log d_j  + \frac{C}{w}\Big).
\end{align}

Since $w=\frac{1}{N} < 1$ and $0 < d_j < 1$, then again $d_j \log w < 0$. Therefore:
\begin{align}
    & \frac{1}{\sum_j d_j} \Big( \sum_j d_j \log w  + d_j \log d_j + \frac{C}{w}\Big) \\
    & \leq \frac{1}{\sum_j d_j} \sum_j d_j \log d_j + \frac{C}{w}.
\end{align}

Since $d_{ii}=1$ then $\sum_j d_j \geq 1$. Therefore:
\begin{align}\nonumber
    & \frac{1}{\sum_j d_j} \sum_j d_j \log d_j + \frac{C}{w} \\\nonumber
    & \leq \sum_j d_j \log d_j + \frac{C}{w} = \frac{C}{w} - \sum_j e^{-D(\pi_i || \pi_j)} D(\pi_i || \pi_j)
    \\\nonumber
    & \leq \frac{C}{w} - \sum_j e^{-D(\pi_i || \pi_j)}
    = \frac{C}{w} - \sum_j e^{\gH(\pi_i)-\gH(\pi_i, \pi_j)}.
\end{align}

Since $\gH(\pi) \geq 0$, we continue to get that:
\begin{align}\nonumber
    & \frac{C}{w} - \sum_j e^{\gH(\pi_i)-\gH(\pi_i, \pi_j)}
    \leq \frac{C}{w} - \sum_j e^{-\gH(\pi_i, \pi_j)}
\end{align}
We can write our lower bound as :

\begin{align}\label{eq:entropy-pairwise-estimator-lower}
    \hat{\gH}_D(\pi) \geq \gH(\pi|W) +\underset{i}{\sum}w_i \sum_j e^{-\gH(\pi_i, \pi_j)}
\end{align}

where we have omitted the constant $\frac{C}{w}$ since it does not depend on $\pi$, and thus does not alter the optimization problem. 

We finally arrive at the joint optimization function:

\begin{align}\nonumber
J(\pi) =
\sum_{t=0}^T \sum_{i=1}^N \, \mathbb{E}_{s_t,a_t \sim \rho_{\pi_i}} &
\big[ r(s_t,a_t) + \alpha_i \gH(\pi_i(\cdot | s_t)) 
\\\nonumber
 + \sum_j & \beta_{i,j} e^{-\gH(\pi_i(\cdot | s_t), \pi_j(\cdot | s_t))} \big],
\end{align}

\section{Complementary Mixture}
Tuning a parameter for each pair results in a quadratic number of optimization problems, which can be cumbersome for mixtures of many components. In order to reduce the need to optimize for $N^2$ different values, we propose measuring a single distance between each component and its \emph{complementary mixture}. For each index $i$, define the complementary mixture $\pi_{-i}$ as the mixture obtained after omitting $\pi_i$. That is: $\pi_0,...\pi_{i-1}, \pi_{i+1}, ...\pi_N$. The distance between each policy and its complementary mixture is then given by:
\begin{align}\label{eq:reduced-pairwise}
    D(\pi_i || \pi_{-i}) = \sum_{t=0}^T \sum_{j \neq i} \mathbb{E}_{s_t,a_t \sim \rho_{\pi_i}} w_j D(d_{\pi_i,t} || d_{\pi_j,t}).
\end{align}
With this reduction we calculate a single distance measure, $D(\pi_i||\pi_{-i})$, for each mixture member, and tune a single temperature value $\beta_i$. Plugging \Eqref{eq:reduced-pairwise} into the estimator in \Eqref{eq:entropy-pairwise-estimator} we can now write the maximum entropy objective function for mixture policies:
\begin{align}\nonumber
J(\pi) & =
\sum_{t=0}^T \sum_{i=1}^N w_i \, \mathbb{E}_{s_t,a_t \sim \rho_{\pi_i}}
\big[ r(s_t,a_t) + \alpha_i \gH(\pi_i(\cdot | s_t)) \\\label{eq:simplified-mmerl-objective}
& -\beta_i \log \big( w_i + w_{-i} \exp{(-D(d_{\pi_i,t} || d_{\pi_{-i},t})} \big) \big],
\end{align}
where $w_{-i} = \sum_{j \neq i} w_j$, and we used the fact that $D(p_i ||p_i) = 0$.
Examining \Eqref{eq:maximum-mixture-entropy-objective} and \Eqref{eq:simplified-mmerl-objective}, we see that it is now possible to decompose the joint maximum mixture entropy objective function into $N$ marginal optimization problems: 
\begin{align}\label{eq:marginal-mmerl}
J_i(\pi) = 
\sum_{t=0}^T \, \mathbb{E}_{s_t,a_t \sim \rho_{\pi_i}}
\big[ & r(s_t,a_t) + \alpha_i \gH(\pi_i(\cdot | s_t)) \\\nonumber
& +\beta_i  D(d_{\pi_i,t} || d_{\pi_{-i},t}) \big].
\end{align}
Note that we have absorbed the mixing weights $w_i$ into the temperature coefficients and used the fact that $\argmax_x \log(1+x) = \argmax_x \log(x)$.
\par
Calculating the distance $D(\pi_i || \pi_{-i})$ between component $i$ and its complementary mixture $\pi_{-i}$ requires broadcasting the experience that $\pi_i$ collects to the rest of the mixture. However, we recall that increasing the exploration via experience sharing was our original motivation in using mixture policies in the first place.

% %%%%%%%%%%%%%%%%%%%%%%%%%%%%%%%%%%%%%%%%%%%%%%%%%%%%%%%%%%%%%%%%%%%%%%%%%%%%%%%
% %%%%%%%%%%%%%%%%%%%%%%%%%%%%%%%%%%%%%%%%%%%%%%%%%%%%%%%%%%%%%%%%%%%%%%%%%%%%%%%

\end{document}